# Finite Element Analysis and Machine Learning Guided Design of Carbon Fiber Organosheet-based Battery Enclosures for Crashworthiness

Shadab Anwar Shaikh[1], M.F.N. Taufique[1], Kranthi, Balusu[1], Shank S. Kulkarni[1,2], Forrest Hale[1], Jonathan Oleson[3], Ram Devanathan[1], Ayoub Soulami[1]

1 Pacific Northwest National Laboratory, Richland, WA, USA
2 The University of Tennessee Knoxville, TN, USA
3 Michigan Technological University, Houghton, MI, USA

Abstract:

Carbon fiber composite can be a potential candidate for replacing metal-based battery enclosures of current electric vehicles (E.V.s) owing to its better strength-to-weight ratio and corrosion resistance. However, the strength of carbon fiber-based structures depends on several parameters that should be carefully chosen. In this work, we implemented high throughput finite element analysis (FEA) based thermoforming simulation to virtually manufacture the battery enclosure using different design and processing parameters. Subsequently, we performed virtual crash simulations to mimic a side pole crash to evaluate the crashworthiness of the battery enclosures. This high throughput crash simulation dataset was utilized to build predictive models to understand the crashworthiness of an unknown set. Our machine learning (ML) models showed excellent performance ($R^2$ > 0.97) in predicting the crashworthiness metrics, i.e., crush load efficiency, absorbed energy, intrusion, and maximum deceleration during a crash. We believe that this FEA-ML work framework will be helpful in down select process parameters for carbon fiber-based component design and can be transferrable to other manufacturing technologies.

## 1 Introduction

Industry and government agencies worldwide are working to reduce reliance on fossil fuels powered vehicles because of their adverse environmental impacts, unpredictable fuel prices, and limited fuel supply. As a result of these organizational efforts, there has been increased adoption of electric vehicles (E.V.s) in the past few years, and it is expected to rise further [1]. Although E.V.s offer a promising alternative to petroleum-powered vehicles, at least two issues can prevent the full realization of their potential. The first issue would be the heavier weight of E.V.s. They are heavier because of the battery packs and the associated battery enclosure, reinforced framework, and suspension [2]. The second issue would be fire safety, if the battery pack is damaged, there is a large risk of battery thermal runaway, fire, and explosion [3].

One way to reduce the weight of E.V.s is to use carbon fiber reinforced polymer (CFRP) composites for the battery pack's enclosure/box. The battery enclosure is typically made of aluminum-based high-strength alloys [4]. CFRP composites, in comparison, offer specific strengths and stiffnesses that are a few times higher. For this reason, aircraft components that are traditionally manufactured using aluminum alloys are being replaced by CFRP composites, and now composites make up most of the modern aircraft's materials [5]. Similarly, the use of composites in the automotive sector continues to grow rapidly[6]. For these reasons, battery enclosure designs using composites have already been explored [7]. For instance, it has been found that a CFRP composite with a comparable strength would provide significant weight savings[8].

Considering only the weight savings, the choice of using CFRP for the battery enclosure is justified. However, fire safety is a critical issue that must be considered before CFRP battery enclosure can be qualified to be widely deployed. CFRP materials by themselves can be fire and corrosion resistant [9]. However, fire originates not in the enclosure but from the battery pack because of the damage to it [3]. Therefore, the role of enclosure in overall fire safety must be evaluated with respect to how well it can prevent damage to the battery pack inside. The battery enclosure is typically mounted at the bottom of E.V.s, as a vehicle is more prone to rear and frontal impacts in the event of an accident. Nonetheless, the battery enclosure can still be damaged by ground impacts [10] and impact to the side during crashes [3]. The later impact mode is of interest to this study. Damage to just the battery pack has been extensively studied, and the kind of impacts that lead to a fire are well understood[11] – [13]. The damage to the battery pack when within an enclosure during a crash was much less studied. In a particular study, the design of the battery pack was optimized for crashworthiness, but the load-bearing components of this enclosure were all made of steel [14].

The intent to use CFRPs for crashworthy battery enclosures presents both opportunities and challenges. Designing and manufacturing a CRFP enclosure would involve setting many design and process parameters. Design parameters would be the number of layers, the orientation of fibers, and the stacking arrangement. Note that parameters governing the shape of the enclosure also need to be set. However, this aspect is skipped in this study because the same problem is encountered for metallic enclosures, too, and is not unique to CFRP enclosures. Processing parameters during thermoforming would mean the velocity of the punch during forming and the temperatures during the process. In total, there are many parameters, and this could present an opportunity to tailor the properties that would lead to a highly efficient crashworthy enclosure. The challenge would be knowing the effect of changing each of these parameters on crashworthiness and determining the optimal values. The challenge would be amplified by the fact that crashworthiness is not a single parameter because commercial safety standards like those for cars do not exist for battery enclosures.

Optimizing many parameters by extensive experimentation and trial-and-error can be extremely challenging, laborious, and inefficient. Therefore, physics-based simulation tools followed by data-intensive machine learning models must be relied on to predict crashworthiness for all possible parameter combinations. The simulations required here could be classified under the process-property modeling task often encountered in Integrated Computational Materials Engineering (ICME) [17]. The simulation task can be broken down into three steps. The first step would be process-structure modeling, where the thermoforming process is simulated for all parameter sets to determine fiber orientations in the formed enclosure. For this purpose, ESI Inc. has a commercially available software PAM-FORM[18], that uses the finite element method (FEM) to simulate the thermoforming process used to manufacture carbon fiber composite [19]. Such software is ideal because it can consider the mechanics of fiber matrix interaction during thermoforming [20]. Moreover, this software can be utilized to perform high throughput thermoforming simulations by varying the process parameters related to composite materials. The second step would be structure-property modeling, where all enclosures' side crashes are simulated. For this purpose, virtual performance simulation (VPS)[21] software offered by ESI Inc. is suitable because it can capture complex damage behaviors of composites during crashes [22]. While there are other approaches for structure-property modeling in composite [23]– [25], using VPS in conjunction with PAM-FORM would make building the pipeline to transfer forming results to a crash simulation easier. By completing steps one and two, a large dataset related to carbon fiber-based battery enclosures is varied by processing parameters, resulting in crash data. This dataset can potentially be utilized in decision-making for unknown/new sets of experimental design spaces by employing artificial intelligence and machine learning. Therefore, in step three, we utilized the dataset generated in steps one and two to predict the crashworthiness of the carbon fiber-based battery enclosures. The predictive accuracy of the machine learning models is highly accurate, and the use of these data-intensive techniques on composite manufacturing technologies holds much promise[26]– [29].

This paper is organized as follows. Section 2 first describes the data generation part of this study, i.e., the methodology used to simulate thermoforming and crash, the selection of the design space, and the parameters extracted to evaluate crashworthiness. Subsequently, section 2 describes the machine learning model selection. Next, section 3 describes some of the results from thermoforming and crash simulation and then, the predictive accuracy of the machine learning models.

## 2 Methodology

In this section, we share the details of our approach for this study. We start by delineating the data generation step. Later we sum up this section by shedding some light on the machine learning model used.

### 2.1 Data Generation

The most critical aspect of this study was the data generation process that involved automating the simulation chain. The results from these simulations were later post-processed to obtain the final crash parameter values. PAM-FORM and VPS were used to perform the thermoforming and crash simulations, respectively.

Although the data generation step involved performing many simulation evaluations with various design parameters, to be consistent, we discuss the simulation setup of a specific case first, followed by automation and later post-processing. The following sections explain these in more detail.

#### 2.1.1 Simulation Setup

The simulation chain shown in Figure 1 is a two-step process, the first involving thermoforming simulation followed by a side pole impact crash simulation. In addition, this also involved pipelining the successfully formed part, i.e., draping, the results from thermoforming into the crash simulation.

##### 2.1.1.1 Geometry of battery enclosure

The battery enclosure geometry Figure 2a, b used in this study was based on the overall dimensions of the E.V. battery pack [30], [31] [48]. However, several simplifications were added. The enclosure was assumed to have no internal or external mounting features. Further, no separate battery module was considered; instead, a nonstructural mass was added to the simulations to mimic the battery module. In addition, a large relief angle was provided to the ends of the enclosure to facilitate easy removal and prevent jamming while manufacturing it using thermoforming simulations. Furthermore, the enclosure was reinforced with ribs to provide additional support, which also served as a demarcation between different battery modules. Finally, the whole enclosure was covered with a lid. The lids, and ribs were assumed to be made of elasto-plastic material.

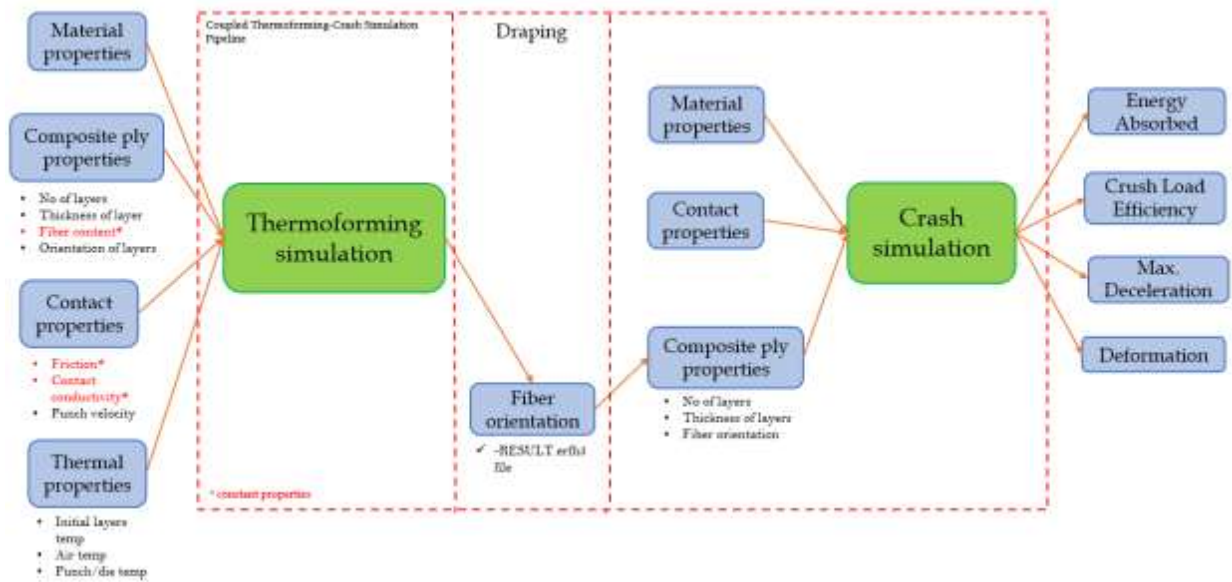

*Figure 1 Simulation chain*

*2.1.1.2   Thermoforming simulation of carbon fiber organosheets*

The schematic demonstrating the thermoforming simulation process is shown in Figure 3 [48]. The material properties used for the punch, die, and organosheet thermal are listed in in the appendix. Initially, an unformed multi-layered carbon fiber organosheet was placed between a punch and a die. The initial temperature of the die and punch was kept at 153 °C, whereas the carbon fiber sheet was heated to 255 °C. Furthermore, the ambient temperature was maintained at 18°C. The organosheet was meshed within the software using 2D shell elements with a mesh size of about 3 mm. The gap between the punch and a die was kept at 1 mm, given the fact that four organosheet, each 0.25 mm thick, were used. However, this would change if a different count and thickness were to be used. The contact between the punch and layers and in-between the layers were considered with friction penalty and thermal conductivity of 0.45 W/m.K.

The simulation starts with a downward punch stroke along the Z-direction with a velocity of 5 m/s while the die is kept stationary. During the downward stroke of the punch, the organosheet is pressed against the die, which plastically deforms the organosheet; this is continued further until the forming process is complete. A 15-degree relief angle is provided on both the punch and die to facilitate the easy removal of the punch after stroke completion,

To alleviate convergence issues during simulations that arise because of jamming of organosheets due to excessive wrinkling during the forming process, a square cut is provided at each corner of the sheet. The dimension of the square can be calculated using a simple relationship:

$$Size\ of\ square\ cutout = \sqrt{\frac{Outer\ surface\ area\ of\ punch - Area\ of\ organosheet}{4}} \quad (1)$$

It was found that for a sheet size of 1500 × 1052 mm, the size of square cutout was about 105 × 105 mm didn't cause any convergence issues in any simulations.

*2.1.1.3    Crash simulation of composites.*

The side pole impact test is a standard test suggested by European New Car Assessment Program[32] and USA's National Highway Traffic Safety Administration (NHTSA)[33] and is the "worst case" scenario to test the crashworthiness of battery enclosure. In this step we perform a side pole impact test of a battery enclosure using VPS software based on FEA. The whole enclosure module is loaded with a nonstructural mass of 100 Kg to mimic the battery module and is covered with a lid. The material properties of ribs and lid is shown in the Figure 4 and are connected to the enclosure with a tied contact; furthermore, during the simulation the lid is restricted to move in the vertical Z direction. In addition, the rib and lids are modeled as an isotropic elasto-plastic material. The enclosure is modelled using bidirectional elasto-plastic material with progressive damage[34], the material properties are shown in the appendix, and is meshed with a 2D shell element with a mesh size of 5 mm.

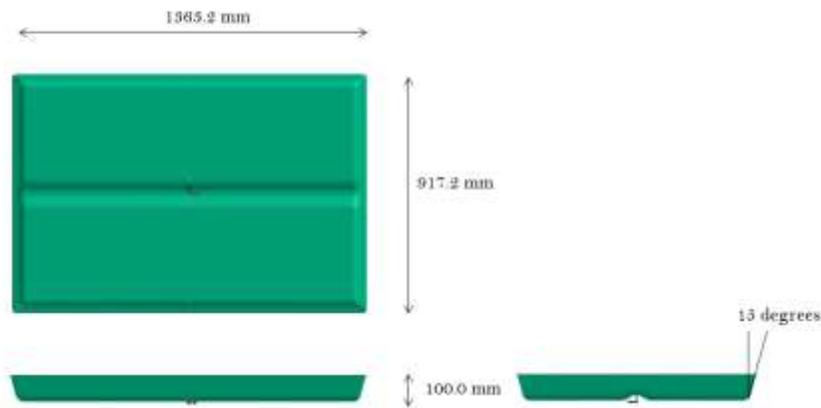

*Figure 2a Geometry of the battery enclosure*

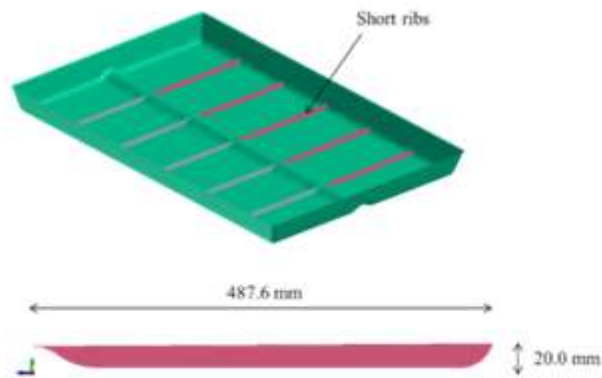

*Figure 2b Placement of ribs inside the battery enclosure and front view of a single rib (Taken from previous publication under same project* [48]*)*

In thermoforming simulation, due to shearing, the fiber orientation at certain location of the composite sheet gets severely distorted. This can influence the overall performance of the formed part. To capture these changes in fiber orientation in crash simulations the enclosure geometry is draped with the results from the thermoforming simulations. The draping process updates the material law of the enclosure material model to reflect these variations in the fiber orientation.

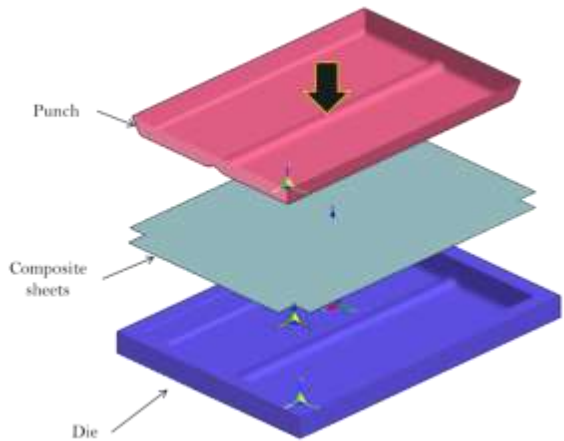

*Figure 1 Thermoforming simulation setup (Taken from previous publication under same project [48])*

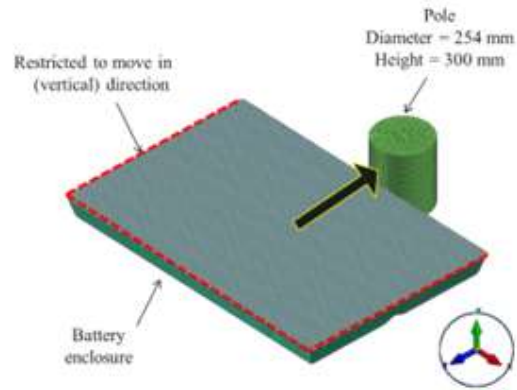

*Figure 2 Crash simulation setup (Taken from previous publication under same project [48])*

After successfully draping, the battery enclosure is impacted on the stationary rigid pole with a velocity of 35 km/hr. for 10 milliseconds as shown in the Figure 2. The contact between pole and enclosure was defined with a contact thickness equivalent to half of the laminate thickness, coefficient of friction of 0.2, and stiffness proportional damping of 0.01. The output of the simulation was saved after every 0.01 milliseconds.

### 2.1.2 Automation

The simulation files were generated by automating and executing the thermoforming-crash simulation chain, shown in Figure 1, on points sampled from the design space. The design space consisted of 7 variables that were chosen from the list of composite ply, contact and thermal properties. The sample points were generated by building a design of experiment (DOE) matrix using Latin Hypercube (L.H.) sampling technique [35] on the design space for 400 samples. Various design properties and ranges that were considered for creating DOE matrix is shown in Table 1. Later, there files were run on the high-performance computing (HPC) cluster.

*Table 1 Range of different properties used for L.H. sampling.*

| Properties | Range |
| --- | --- |

| Number of composite layers | 4 - 16, only even |
|---|---|
| Thickness of each composite layers | 0.1 - 0.6 mm |
| Fiber orientations | (0, 45, -45, 90), (30, -30, 60, -60) |
| Punch velocity | 4 - 6.5 m/s |
| Layer temperature | 200 - 400 °C |
| Punch / Die temperature | 20 - 220 °C |
| Air temperature | 10 - 30 °C |

PAM-FORM was automated to take the DOE matrix as an input and generate simulation data files, and these files were run on a high-performance computing cluster (HPC) cluster. Likewise, VPS was automated to drape the results from thermoforming and generate the crash simulation files. Further, the crash simulations were performed, and results were processed to obtain final crash parameter values. Both PAM-FORM and VPS were automated using Python.

In total, 400 thermoforming simulations were performed; out of these, about 65% of the simulations were successful. This is because, for a given set of design and process parameters, it is not possible to manufacture a part using thermoforming. The results of these simulations were later used to perform crash simulations. Each thermoforming and crash simulation was run on 16 and 8 cores, respectively. Further, the simulations with increasing layers were found to be computationally more expensive. A 10-layer thermoforming simulation, for example, took around 16 hours, and the crash took about 45 minutes on the above-mentioned computing units.

### 2.1.3   Post-Processing

Four parameters were extracted from the simulation results to assess the enclosure's crashworthiness. These parameters are crush load efficiency (CLE), energy absorbed (E.A.), intrusion, and deceleration.

#### 2.1.3.1   Crush load efficiency (CLE)
The ratio of the average net force on the enclosure to the maximum net force on the enclosure is known as crush load efficiency (CLE). This parameter is a good indicator of the stability of the crushing process [36] and greatly influences the head injury criteria (HIC) used in passenger safety testing. The higher value of CLE implies a lower impact on the internal contents of the enclosure and, therefore, better crashworthiness. The theoretical maximum value of CLE is 1.

#### 2.1.3.2   Energy absorbed (EA)
The energy absorbed by the enclosure during the crash, i.e., the energy absorbed (E.A.), is a crucial indicator of crashworthiness. During the crash, some kinetic energy is dissipated as heat through material deformation or damage. Higher dissipated energy implies lower kinetic energy remains, thereby decreasing the chances of damaging the internal contents of the enclosure. Some car structures, termed crumple zones, are specifically made to promote E.A. [37]. Its value is either the decrease in the kinetic energy of the enclosure after the impact or the increase in the internal energy of the enclosure. E.A. values were expressed in Joules.

*2.1.3.3 Intrusion*

The maximum intrusion of the battery enclosure directly relates to the safety of the battery packs inside. Its importance is reflected in the fact that there are extensive safety studies on thermal runaways and fires resulting from the intrusion of rigid objects into battery cells were conducted [3]. A higher intrusion value would imply higher chances of battery fire and, therefore, lower crashworthiness. The intrusion value in millimeters (mm) includes the deformation due to plastic and elastic deformations.)

*2.1.3.4 Deceleration*

During the crash, the deceleration on the battery pack might be high even if there is no direct mechanical impact[38]. These deceleration forces might lead to an external short circuit that causes a fire [3]. Therefore, the maximum deceleration the enclosure, including the battery pack, undergoes during the crash indicates crashworthiness. A larger deceleration value would mean that the enclosure is less crashworthy. Deceleration values were expressed in $m/s^2$.

The choice of these parameters was made through a review of the crashworthiness standards of cars and crash studies of battery packs. Additional parameters that could help assess the crashworthiness could be the specific energy absorbed (SEA) and maximum force. The former can be found by dividing E.A. by the mass of the enclosure and the latter by multiplying maximum acceleration with the mass. Mass, in turn, is a known function of the number of layers and thickness of the layers. Therefore, consideration of these two parameters as the machine learning models' outputs was unnecessary.

## 2.2 Machine Learning Models

### 2.2.1 Data Collection and Feature Generation

The crash simulation generated a total of 266 individual crash tests. Each of the crash samples indicates a unique data point in the design space, which includes seven parameters. These are the number of composite layers, the thickness of layers, fiber orientations, punch velocity, the initial temperature of layers, punch temperature, and air temperature, as presented in Table 5. We call these parameters features for the machine learning model. After the post-processing, four crucial parameters related to the crash test were calculated from the crash test. These parameters are CLE, EA, intrusion, and deceleration. We call it target properties. Once all the features and corresponding target properties for all the crash samples were collected, the train-test split was applied randomly to separate 80 percent of the data as training and 20 percent as testing for the machine learning model. Therefore, the total number of training samples was 212, and the testing sample was 54.

### 2.2.2 Machine Learning Model Construction

Three tree-based ensemble methods were used to predict the target properties from the crash simulation, i.e., Gradient Boosting, Extreme Gradient Boost (or Xgboost), and Random Forest (R.F.). Tree-based ensemble models have demonstrated superior performance compared to other models for predicting materials properties[26]– [29], [39]. Ensemble methods are meta-algorithms that combine multiple base models to create a more accurate predictive model. Bagging ensemble methods can reduce variance, while boosting ensemble methods can reduce bias. Boosting methods can convert weak learners into strong ones, with decision stumps typically being used as base weak learners, though not always[40]– [42]. These methods build models stage-wise and generalize the model by optimizing an arbitrary differentiable loss function. Boosting techniques also help to prevent over-fitting, handle nonlinear relationships between target properties and features, and manage feature collinearity. Additionally, most boosting methods provide information on feature importance, which is crucial for determining which features greatly influence target properties. Tree-based models perform better compared to other models, such as neural networks, support vector machine (SVM), and logistic regression for the case of the smaller dataset in the materials science domain reported by several researchers[28], [29].

The grid-search algorithm was employed to minimize errors by optimizing hyper-parameters. This involved testing all possible combinations of the supplied hyper-parameter values and selecting the one with the least error for our model. It is important to note that each algorithm has a unique set of hyper-parameters. Once the best hyper-parameters were identified for each target property, we used the optimized model to predict the target properties of our test set, which had been unseen to the model. To conduct our study, we utilized the scikit-learn machine learning library for the Python language to employ all ML models [43], including the Xgboost model that was implemented using the library developed by Tianqi Chen [44]

# 3 Results and Discussion
## 3.1 Thermoforming Simulation

The simulations were performed based on the simulation setup discussed in the previous sections. The results obtained are shown in Figure 3(a)-(d). As observed in (a), the stress did not exceed the failure limit of the fibers; hence, no fiber breakage was observed. Further, the organosheet was able to take the shape of the die perfectly without any wrinkles or wrappage. In addition, the surface finish was good, and all the curved surfaces of the die were formed accurately.

During the thermoforming process, the organosheet undergoes severe deformation. As a result, the fiber angle changes from its original value of 90 degrees. The new fiber orientations are shown in Figure 3(b) and (d). It can be observed that, for the most part, fiber orientation stays the same except at several places involving curvature where it gets severely distorted, as seen in Figure 3(d). This new fiber orientation influences the overall mechanical properties of the formed part. Hence these new orientations were mapped to the battery enclosure part in the crash simulation discussed in the next section.

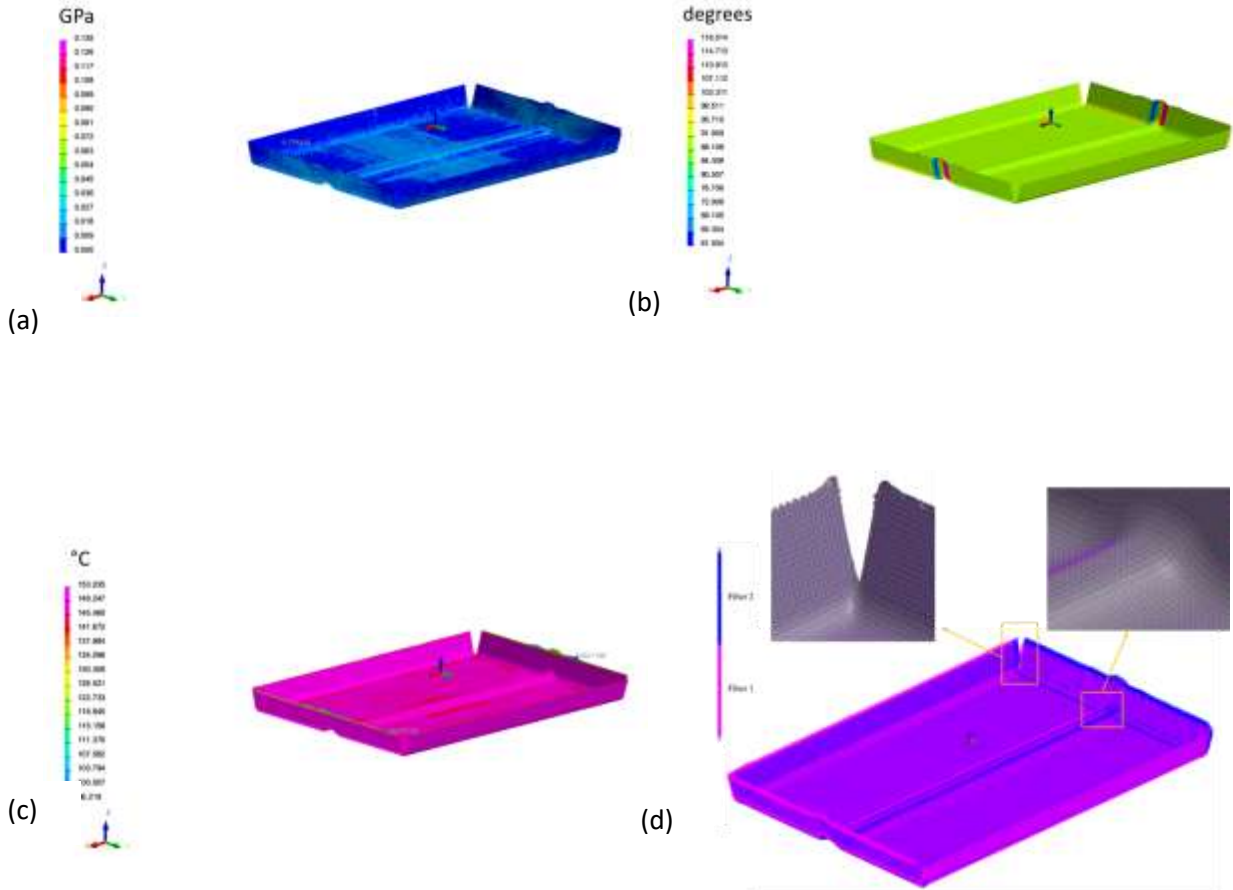

*Figure 3 (a) Total stress distribution, (b) Fiber angle, (c) Temperature distribution (d) Fiber orientations of the formed part*

### 3.2 Crash Simulation

Crash simulations were performed to simulate the side pole impact test of the formed enclosure with explicit dynamic analysis. Due to large deformation with the nonlinear response, the explicit time marching method was employed. It was observed that the battery enclosure assembly bounces back after the impact with a rigid pole without substantial permanent deformation. Figure 4Figure 4 shows the (a) Von Mises's stress and (b) strain distribution at the peak of the impact. High-stress concentration was observed at the point of impact. Figure 4 (b) shows the maximum and average contact force value observed during impact. These values were further utilized to calculate other parameters. Figure 4 (d) shows the energy interaction during the impact. It can be seen that a rise in internal energy accompanies a reduction in kinetic energy. However, the total energy during impact was observed to be constant.

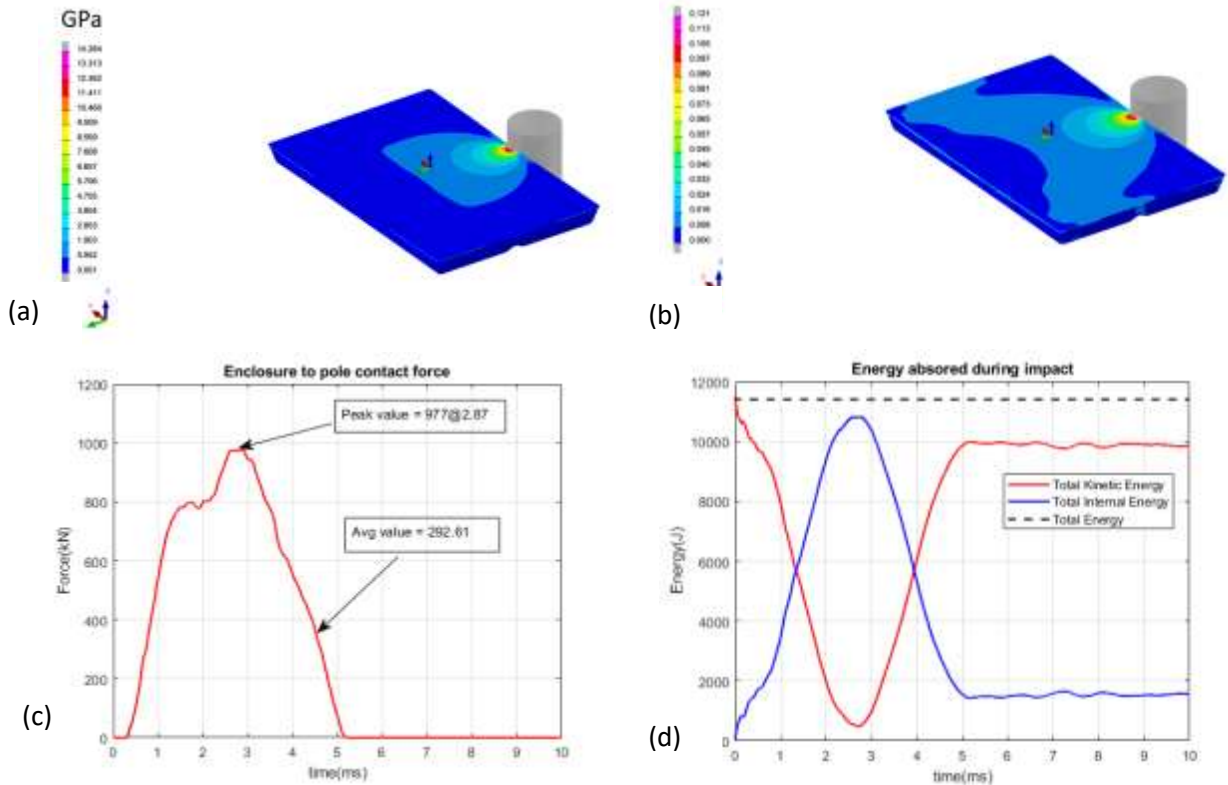

*Figure 4 (a)Von-mises stress and (b) strain distribution, (c) contact force (d) energy absorbed during enclosure impact.*

### 3.3 Predictions from Machine Learning

To begin with, the ML models were trained individually for four target properties, and their hyper-parameters were optimized. Subsequently, we calculated the training and testing error using five-fold cross-validation. We then used these hyper-parameters to predict the target properties of the testing data set. Table 6 shows the cross-validated mean absolute error (MAE) values and mean absolute percentage error (MAPE) on the test set for all the models and corresponding target properties. It is evident from Table 6 that the model performance for Gradient Boosting and Xgboost are nearly identical for CLE, intrusion, and deceleration. Only in the case of EA and deceleration, Xgboost performs better than other models in accuracy (i.e., lower MAE). Therefore, due to its excellent performance, we will discuss the feature importance and predicted results generated by the Xgboost model.

*Table 2: Cross-validated MAE and MAPE scores for all the target properties.*

| Model | CLE | | EA | | Intrusion | | Deceleration | |
|---|---|---|---|---|---|---|---|---|
| | MAE | MAPE | MAE | MAPE | MAE | MAPE | MAE | MAPE |
| XGB | 0.0035 | 0.58 | 115 | 2.99 | 0.094 | 0.50 | 215 | 1.72 |
| GB | 0.0033 | 0.54 | 227 | 6.65 | 0.085 | 0.46 | 447 | 2.62 |
| RF | 0.0041 | 0.67 | 688 | 11.10 | 0.109 | 0.58 | 458 | 3.52 |

### 3.3.1 Feature Importance

Following the training of models for all the target properties using optimized hyper-parameters, we evaluated the feature importance of the Xgboost model. Feature importance is a score assigned to features based on their usefulness in predicting a target variable. Figure 7 presents the feature importance for all the target properties. The figure shows that the number of composite layers and organosheet thickness are the two most important factors determining all four crashworthiness parameters. However, for CLE, the influence of thickness is much lower than that of the number of layers. It should be noted that the air temperature during thermoforming also has a nonnegligible influence on deceleration.

The prediction that the number of layers and organosheet thickness are the two most influential factors on crashworthiness makes sense. Increasing the number of layers and thickness means more material to spread the impact energy, resulting in a higher amount of absorbed energy (EA). Increasing the number of layers and thickness increases the enclosure's strength and bending stiffness [11,12]. Increasing strength and stiffness would imply lower intrusion and higher deceleration.

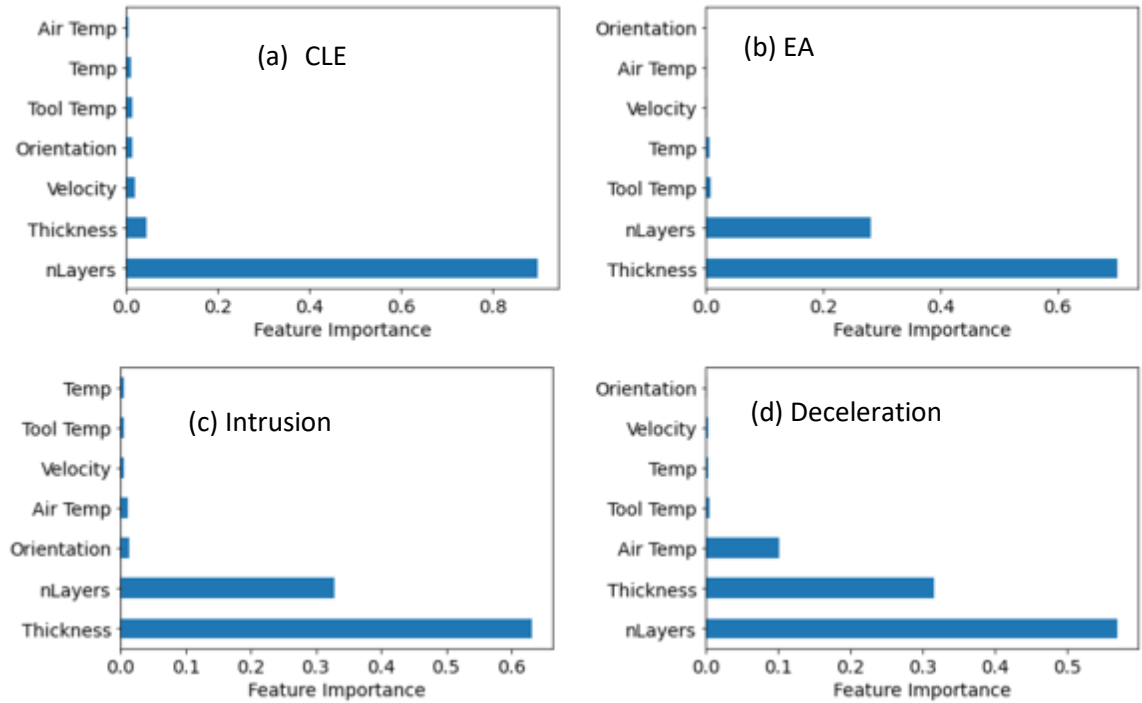

*Figure 7: Feature importance calculated by the Xgboost method for different target properties.*

### 3.3.2 Prediction for the Test Data

We finally used the trained Xgboost model to predict the four target properties of the composites. A comparative plot for the FEA-measured target properties and Xgboost-predicted target properties is presented in Figure 8 (a to e). In this plot, we measured the coefficient of determination ($R^2$) for all the target properties to make predictive accuracy more generalizable since $R^2$ is independent of scale. Figure 8 shows that the predictive accuracy for the Xgboost model for CLE, E.A., intrusion, and deceleration is excellent in terms of $R^2$ values. In fact, for these target properties, the predictive capability is well for both lower- and upper-bound data, which signifies that the model learns from the data patterns even though the training data size was limited. We believe the model might be used as a swift screening tool to down select the process parameter spaces from large DOE spaces.

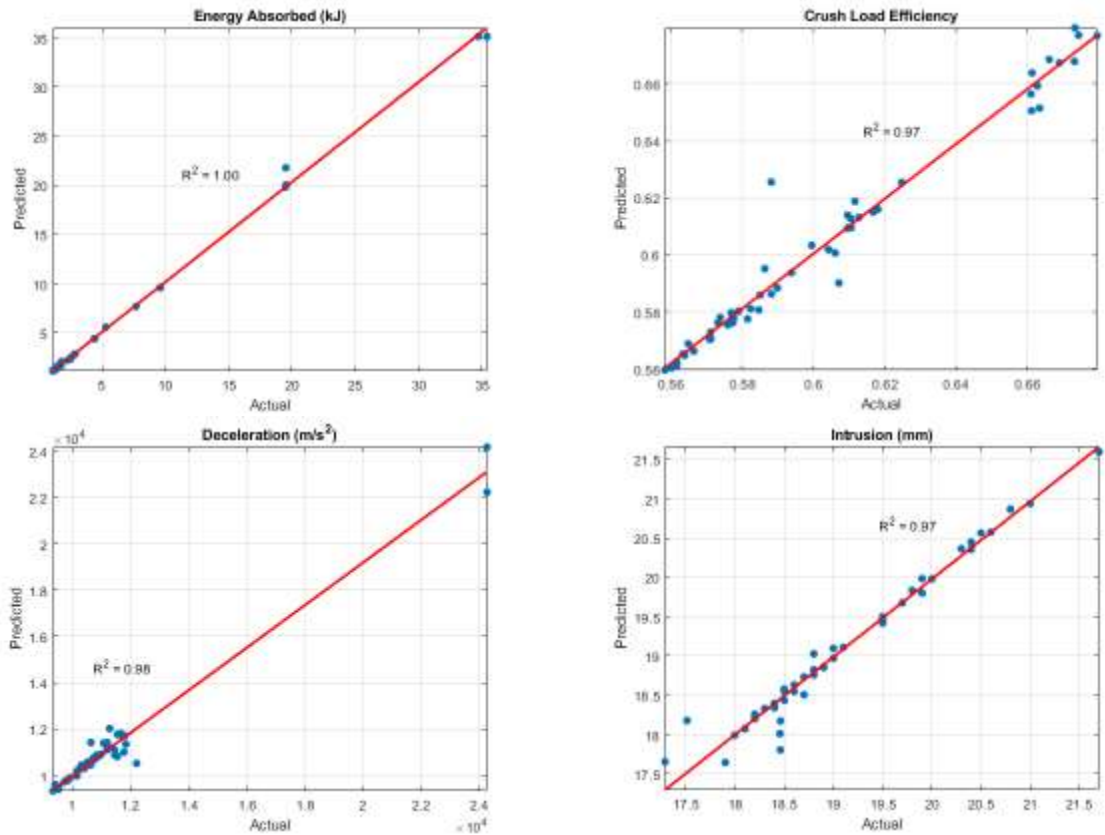

*Figure 8: Comparison between the simulation and Xgboost predicted target properties.*

### 3.3.3  Symbolic regression

The above-mentioned ML models provide accurate predictions of the target properties with feature importance, which can shed light on the important parameters for composite design. However, these models still fail to make a functional relationship between the input features and output target properties and hence have poor generalizability. To solve this issue, we employed Symbolic regression, a special type of ML that focuses on finding a mathematical expression that fits a given dataset. It can be seen as a form of "equation discovery," where the goal is to find a compact and interpretable equation that accurately represents the data. To implement symbolic regression, only the dataset corresponding to "type B" orientation was considered since it has 186 data points out of 266. Like Xgboost models, for symbolic regression, 80 percent of the data (138 data points) was considered as testing, and 20 percent of the data (38 data points) was considered as a holdout test set. We used commercially available symbolic regression tools called TuringBot[45], which is highly engineered and robust, with broad applicability in many domains[46], [47] the symbolic regression analysis. The symbolic regression generated equation and corresponding performance metric, i.e., $R^2$ and MAE, is presented in Table 7. Figure 9 provides the comparison between the actual values and symbolic regression predicted values.

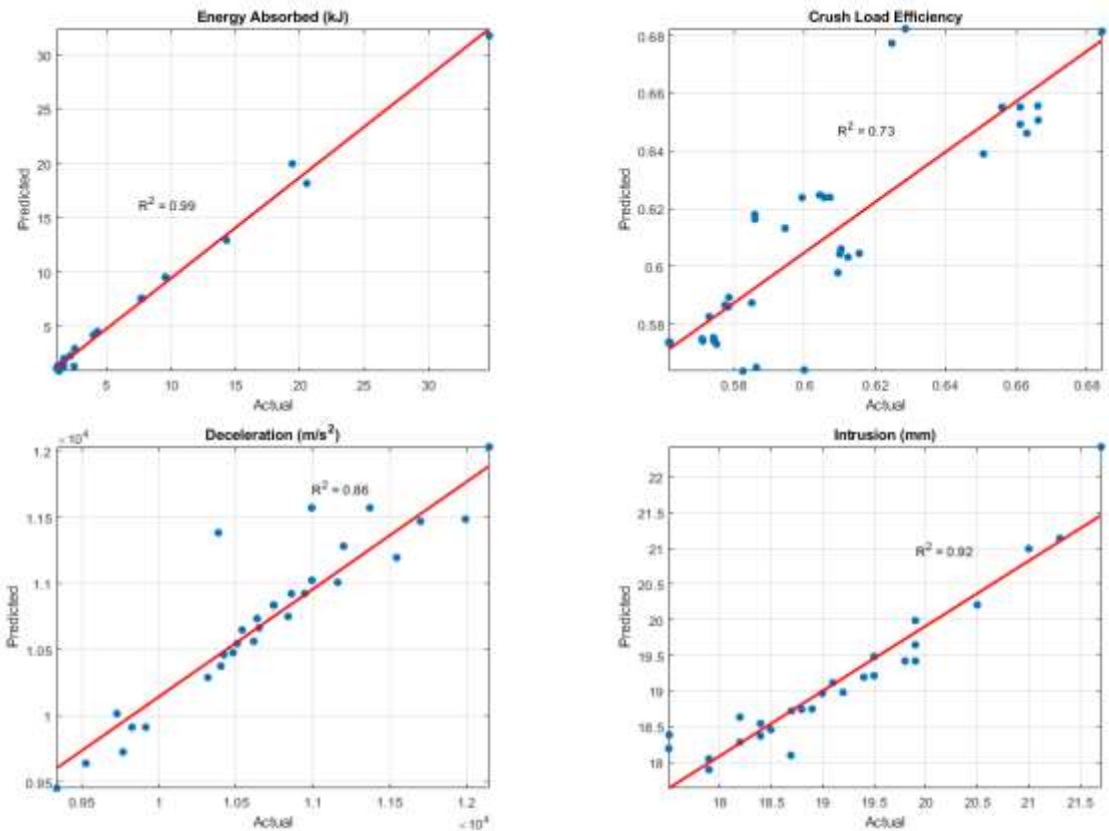

*Figure 9: Comparison between the simulation and symbolic regression predicted target properties.*

From Table 3, it is clear that while generating the equation, it considered the most important features from the Xgboost model presented in Figure 7. These equations are also mainly governed by the number of layers and thickness of the organosheet. However, all the features were considered during the training of the symbolic regression model. It is essential to mention that the nature of the equation will change depending on the dataset. Therefore, these equations and resulting coefficients only apply to this specific dataset. Moreover, the symbolic regression's accuracy is lower than the Xgboost model, as presented in Figure 9 and Figure 8. Usually, traditional machine learning methods such as decision trees can often provide higher accuracy on complex datasets. On a different note, symbolic regression can be helpful in cases where the goal is not just to make predictions but to gain a deeper understanding of the underlying processes that generate the data. Based on our dataset and results, we can infer that the traditional machine learning models followed by symbolic regression will be helpful for better prediction and understanding of the functional relationships among the input features and output target properties.

*Table 3: Symbolic regression generated equation and resulting $R^2$ MAE and MAPE scores for the test dataset. In the equations, $a$ is the number of composite layers, $b$ is the thickness of each composite layer, $c$ is the composite layer temperature, and $d$ is the punch/die temperature.*

| Target property | Symbolic regression generated equation | $R^2$ | MAE | MAPE |
|---|---|---|---|---|
| CLE | $\dfrac{-297024 + 2184c + 273d - 272a^2 + 2ca^2}{490(-1088 + 8c + d)}$ | 0.73 | 0.014 | 2.36 |
| EA | $1303 + 8a^2 b^3 (-33 + (3 + a)ba)$ | 0.99 | 354.21 | 9.33 |
| Intrusion | $-0.692ab + 0.390a + 1.819b + 17.116$ | 0.92 | 0.20 | 1.07 |
| Deceleration | $136.585a - 1874.219b + 10032$ | 0.86 | 531.28 | 1.50 |

## 4  Conclusion

We successfully developed a modeling framework that considers design and manufacturing process parameters to predict the crashworthiness of carbon fiber-based composite battery enclosures. The modeling framework consists of a FEA-based simulation pipeline with thermoforming and crash simulations, followed by ML to predict crashworthiness indicators such as crash load efficiency, absorbed energy, intrusion, and deceleration. For all the target parameters, the $R^2$ values for the holdout test set were more than 0.97, which shows the promise of ML models in determining complex relationships in composite design for vehicle technology. Our study also established the role of the number of composite layers and thickness of the sheets as the most prominent features that govern the crashworthiness of the composites. Additionally, symbolic regression analysis provided the governing equations related to the dataset for further generalizability. We believe that this combined FEA-ML framework will be applicable to other manufacturing platforms pertaining to metallic alloys and polymers.

**Author Contributions:**
Conceptualization, R.D, S.K., M.T. A.S.; methodology, S.S., S.K., B.K., F.H., R.D., and A.S; software, S.S., S.K. B.K., F.H., M.T and J.O; formal analysis, S.S, M.T. and B.K.; investigation, S.S., M.T., and B.K.; writing—original draft preparation, S.S., M.T. and B.K.; writing—review and editing, S.S., M.T., B.K. R.D. and A.S.; supervision, A.S., and R.D.; funding acquisition, R.D. All authors have read and agreed to the published version of the manuscript.

**Acknowledgement:**

The authors would like to acknowledge the help from Arnaud Dereims and Ramesh Dwarampudi (ESI North America) regarding ESI software.

**Funding:**

This research was funded by U.S. department of Energy's Office of Energy Efficiency and Renewable Energy under the proposal FP-D-20.1-23733 and ESI North America Inc.

**Data Availability Statement:**

The data that support the findings of this study are available from the corresponding author upon reasonable request.

**Conflicts of Interest:**

The authors declare no conflict of interest.

# 5 Appendix

*Table 4 Material properties for the die and the punch.*

| Properties | Value |
|---|---|
| Mechanical properties | Rigid material |
| Convection coefficient | 10 W/m² K |
| Conductivity | 0.45 W/m.K |

*Table 5 Material properties for the carbon fiber organosheets.*

| Properties | Value |
|---|---|
| Density | 1e-6 kg/mm³ |
| Initial angle between fibers | 90 degrees |
| Thickness | 0.25 mm |
| Fiber content | 0.5 |
| Tension compression stiffness (Fiber 1) | 20 GPa |
| Tension compression stiffness (Fiber 2) | 20 GPa |
| Bending stiffness (Fiber 1) | 0.03 GPa |
| Bending stiffness (Fiber 2) | 0.03 GPa |
| Conductivity | 2.3e-6 kW/mm °C |
| Specific heat | 1150 J/kg °C |
| In plane shear | 2.5e-5 GPa |
| Sheet orientation | (90,45-,45,0) |
| Layer separation stress | 0.005 GPa |
| Convection coefficient | 10 W/m²K |

*Table 6 Material properties for lid and ribs.*

| Properties | Value |
|---|---|
| Density | 1.8e-6 kg/mm³ |
| Young's modulus | 125 GPa |
| Yield stress | 3.5 GPa |
| Poisson's ratio | 0.33 |
| Max plastic strain for element removal | 0.014 |
| Plastic tangent modulus | 8 GPa |

*Table 7 Material properties of enclosure in crash simulation*

| Properties | Value |
|---|---|
| Density | 1.8e-6 kg/mm$^3$ |
| Young's modulus parallel to fiber | 125 GPa |
| Young's modulus perpendicular to fiber | 8 GPa |
| Critical shear damage limit | 0.114 GPa |
| Initial shear damage limit | 0.02 GPa |
| Initial strain of tensile fiber | 0.012 |
| Ultimate strain of tensile fiber | 0.014 |
| Tensile fiber ultimate damage | 0.99 |
| Initial strain compressive fiber | 0.008 |
| Ultimate strain compressive fiber | 0.009 |
| Compressive fiber ultimate damage | 0.99 |
| Initial yield stress | 0.02 GPa |
| Hardening law exponent | 0.64 |
| Hardening law multiplier | 1.3 |
| Shear modulus 1,2 plane | 7 GPa |
| Shear modulus 2,3 plane | 4 GPa |
| Shear modulus 1,3 plane | 4 GPa |
| Poisson's ratio | 0.33 |
| Critical transverse damage limit | 1 |
| Initial transverse damage limit | 0.02 |

[43]  F. Pedregosa FABIANPEDREGOSA *et al.*, "Scikit-learn: Machine Learning in Python Gaël Varoquaux Bertrand Thirion Vincent Dubourg Alexandre Passos PEDREGOSA, VAROQUAUX, GRAMFORT ET AL. Matthieu Perrot," *Journal of Machine Learning Research*, vol. 12, pp. 2825–2830, 2011, Accessed: Apr. 24, 2023. [Online]. Available: http://scikit-learn.sourceforge.net.

[44]  T. Chen, … C. G. sigkdd international conference on knowledge, and undefined 2016, "Xgboost: A scalable tree boosting system," *dl.acm.org*, vol. 13-17-August-2016, pp. 785–794, Aug. 2016, doi: 10.1145/2939672.2939785.

[45]  "Symbolic regression software - TuringBot." https://turingbotsoftware.com/ (accessed Apr. 24, 2023).

[46]  C. Cornelio *et al.*, "AI Descartes: Combining data and theory for derivable scientific discovery," *arxiv.org*, Accessed: Apr. 24, 2023. [Online]. Available: https://arxiv.org/abs/2109.01634

[47]  D. Ashok, J. Scott, S. J. Wetzel, M. Panju, and V. Ganesh, "Logic Guided Genetic Algorithms (Student Abstract)," *Proceedings of the AAAI Conference on Artificial Intelligence*, vol. 35, no. 18, pp. 15753–15754, May 2021, doi: 10.1609/AAAI.V35I18.17873.

[48]  Kulkarni, Shank S., et al. "Investigation of Crashworthiness of Carbon Fiber-Based Electric Vehicle Battery Enclosure Using Finite Element Analysis." *Applied Composite Materials* (2023): 1-27.
22 | P a g e